\documentclass[runningheads]{llncs}
\usepackage{graphicx}
\usepackage{amsmath}
\usepackage{amssymb}
\usepackage{booktabs}
\usepackage{todonotes}
\usepackage{algorithmic}
\usepackage[linesnumbered,ruled,vlined]{algorithm2e}
\usepackage{multirow}
\usepackage{makecell}
\usepackage{pifont}
\usepackage{caption}
\usepackage{subcaption}
\usepackage{todonotes}
\usepackage{hyperref}

\usepackage{tikz}
\usepackage{comment}
\usepackage{amsmath,amssymb} 
\usepackage{color}
\usepackage{float}

\newfloat{figtab}{htb}{fgtb}
\makeatletter
  \newcommand\figcaption{\def\@captype{figure}\caption}
  \newcommand\tabcaption{\def\@captype{table}\caption}
\makeatother

\SetKwInput{KwInput}{Input}                
\SetKwInput{KwOutput}{Output}              

\SetCommentSty{mycommfont}

\begin{document}

\pagestyle{headings}
\mainmatter
\def\ECCVSubNumber{6223}  

\title{Towards General and Efficient Active Learning} 

\titlerunning{Towards General and Efficient Active Learning}
%
\author{Yichen Xie, Masayoshi Tomizuka, Wei Zhan\thanks{Correspondence}}
\authorrunning{Y. Xie et al.}
%
\institute{University of California, Berkeley\\
\email{\{yichen\_xie,tomizuka,wzhan\}@berkeley.edu}}
\maketitle

\begin{abstract}
Active learning selects the most informative samples to exploit limited annotation budgets. Existing work follows a cumbersome pipeline that repeats the time-consuming model training and batch data selection multiple times. In this paper, we challenge this status quo by proposing a novel general and efficient active learning (GEAL) method following our designed new pipeline. Utilizing a publicly available pretrained model, our method selects data from different datasets with a single-pass inference of the same model without extra training or supervision. To capture subtle local information, we propose knowledge clusters extracted from intermediate features. Free from the troublesome batch selection strategy, all data samples are selected in one-shot through a distance-based sampling in the fine-grained knowledge cluster level. This whole process is faster than prior arts by hundreds of times. Extensive experiments verify the effectiveness of our method on object detection, image classification, and semantic segmentation. Our code is publicly available in \href{https://github.com/yichen928/GEAL_active_learning}{https://github.com/yichen928/GEAL\_active\_learning}.

\end{abstract}

\section{Introduction}
\label{sec:intro}

Deep Neural Network (DNN) models have achieved exciting performance in various tasks. Such success is supported by abundant training samples and labels. Despite the availability of some large annotated datasets \cite{deng2009imagenet,ridnik2021imagenet}, current models are still far from saturation \cite{mahajan2018exploring} with respect to the data amount. Unfortunately, labeling a dataset tends to be time-consuming and expensive, especially for dense supervision tasks such as object detection and semantic segmentation, where experts spend up to 90 minutes per image \cite{lin2019block}. As such, how to exploit the limited annotation budget serves as a long-standing problem for the further development of computer vision and machine learning.

\begin{figure}[tb!]
    \centering
    \begin{subfigure}{0.46\linewidth}
        \centering
        \includegraphics[width=\linewidth]{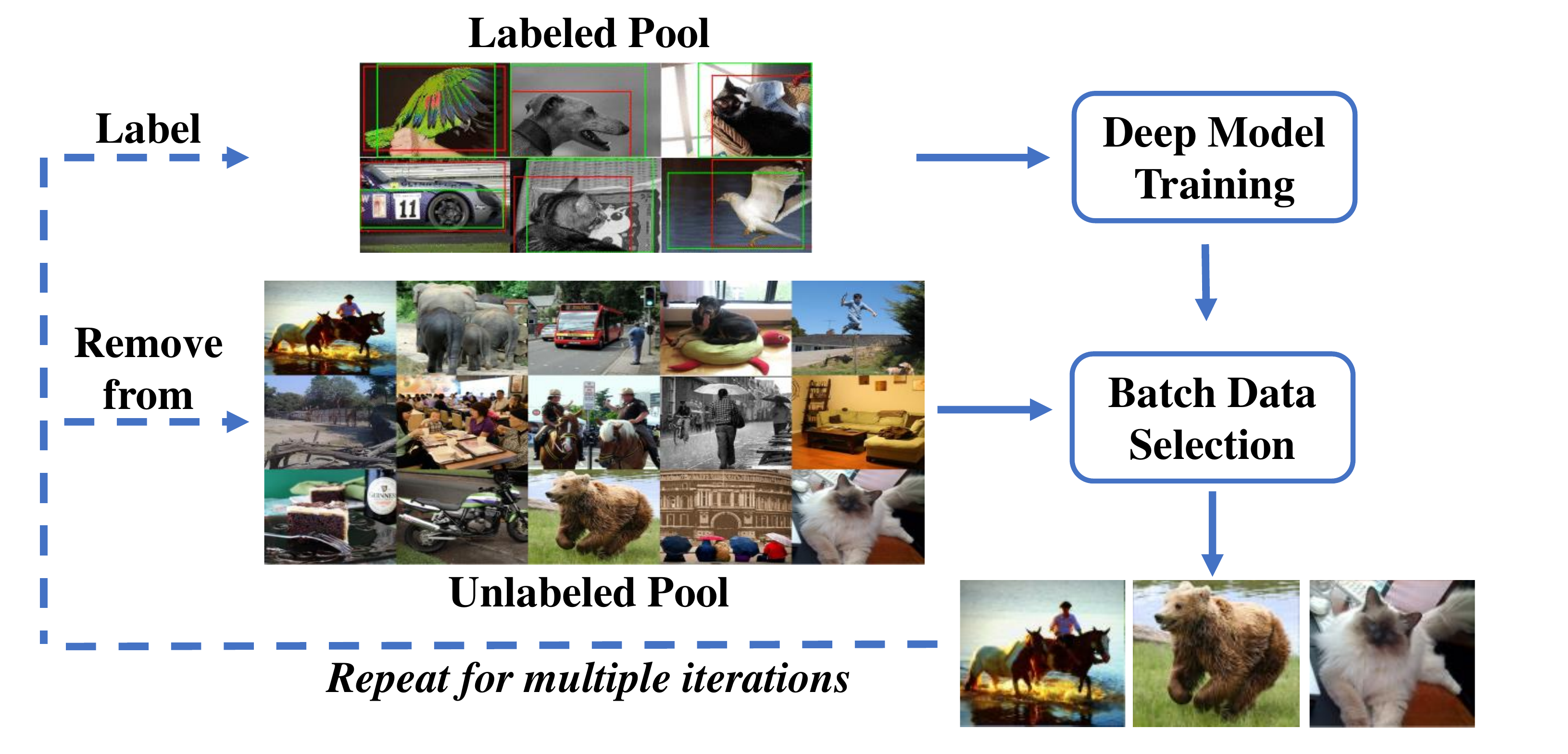}
        \caption{\textbf{Traditional Pipeline}}
        \label{fig:pre_pipeline}
    \end{subfigure}
    \begin{subfigure}{0.53\linewidth}
        \centering
        \includegraphics[width=\linewidth]{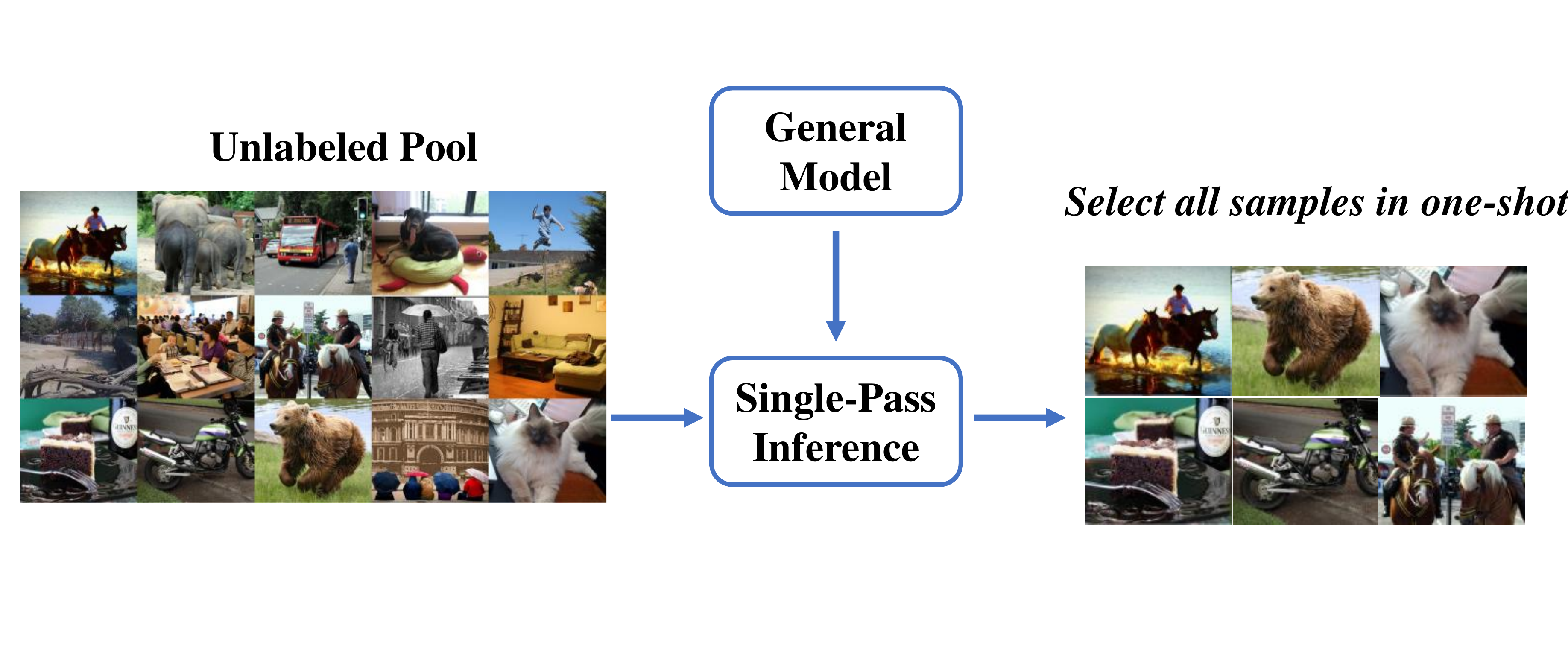}
        \caption{\textbf{Our Proposed Pipeline}}
        \label{fig:our_pipeline}
    \end{subfigure}
    \caption{\textbf{Traditional v.s. Our Proposed Active Learning Pipeline:} In the traditional pipeline, deep models are repetitively trained for multiple times on each dataset. In contrast, our pipeline allows to select the required data samples in one-shot without extra model training.}
\end{figure}

As a potential solution, active learning methods \cite{ren2020survey,settles2009active} attempt to find the most suitable samples for annotation which can best boost the training of task models. In the area of deep learning, almost all existing active learning methods rely on DNNs to process data samples. The costly and time-consuming model training process prevents them from being easily and widely used. To be specific, existing approaches follow a traditional \textit{batch selection} pipeline \cite{sener2017active}, as shown in Fig.~\ref{fig:pre_pipeline}. A deep model is firstly trained on a randomly sampled small initial set, and the model is then used to choose images with certain batch size. These selected images are annotated and put into the labeled pool, and the model is re-trained or finetuned on all the labeled samples again. These steps repeat for multiple iterations on a large dataset, and the whole pipeline should start from scratch when we work on another dataset. In many cases, it even requires up to \textit{several days} to select enough samples from a medium-size dataset (\textit{e.g.} PASCAL VOC \cite{everingham2010pascal} in Tab.~\ref{tab:efficiency}).

In this paper, we challenge this \textit{status quo} by designing a novel active learning pipeline (Fig.~\ref{fig:our_pipeline}). Firstly, we consider the following three principles of our design.

\begin{itemize}
\item \textbf{Generality:} Deep models are necessary for feature extraction in active learning. It is desired that a \textit{general} pretrained model can work on multiple datasets so that it does not require training on each dataset separately.
\item \textbf{Efficiency:} The traditional \textit{batch selection} setting (Fig.~\ref{fig:pre_pipeline}) is known to be time-consuming due to the iterative batch data selection and model training on each dataset. Ideally, it is expected to be replaced with a \textit{single-pass} model inference for feature extraction on each dataset. 
\item \textbf{Non-supervision:} Annotators cannot always respond in time, and the entire active learning progress may be delayed by the frequent requests of labels. This problem exists in many prior work \cite{yoo2019learning,agarwal2020contextual,sener2017active}. To this end, it is preferred that annotations are not required in the pipeline of active learning. 
\end{itemize}

In view of the above principles, we propose a novel \textbf{g}eneral and \textbf{e}fficient \textbf{a}ctive \textbf{l}earning (\textbf{GEAL}) method. To the best of our knowledge, it is the \textit{first} method satisfying all the above principles simultaneously.

\begin{table}[t!]
\caption{\textbf{Generality and Efficiency of Active Learning Methods:} \textit{Train} refers to the necessity of training a deep model separately for each dataset. \textit{Batch} shows whether the method repeats the data selection in batch multiple times. \textit{Supervision} denotes whether it requests ground-truth labels in the data selection process. \textit{Time} estimates the approximate time to select $8000$ images from PASCAL VOC datasets (explanation in Sec.~\ref{sec:analysis}).}
\label{tab:efficiency}
\centering
\begin{tabular}{ccccc}
    \toprule
    \textbf{Methods} & \textbf{Train} & \textbf{Batch} & \textbf{Supervision} & \textbf{Time}\\
    \midrule
    Core-Set \cite{sener2017active} & \ding{51} & \ding{51} & \ding{51} & \multirow{4}{*}{\makecell[c]{$\sim 42\ hours$ \\ +\\ \textit{label query}}}  \\
    Entropy \cite{settles2008analysis,luo2013latent} &\ding{51} & \ding{51} & \ding{51} &\\
    Learn-Loss \cite{yoo2019learning} &\ding{51} & \ding{51} & \ding{51} &\\
    CDAL \cite{agarwal2020contextual}&\ding{51} &\ding{51} &\ding{51} &\\
    \midrule
    GEAL (ours) &\ding{55} &\ding{55} &\ding{55} & $155\ seconds$ ($\sim$975x faster)\\
    \bottomrule
\end{tabular}
\end{table}

The proposed method uses a publicly available pretrained model to extract features from images. After unsupervised pretraining \cite{caron2021emerging} on large scale datasets \cite{deng2009imagenet}, the model bias is small, so we can apply the \textit{same} pretrained model to different datasets. The selected samples are network agnostic, \textit{i.e.} the architectures of pretrained model and downstream task models may be totally different.

In pursuit of \textit{efficiency}, our method gets rid of the repetitive model training in the traditional pipeline. Instead, we directly perform a distance-based sampling algorithm in the level of our newly defined \textit{knowledge clusters} which come from the intermediate features of the pretrained model. The whole data selection process for each dataset is completed with only a \textit{single-pass} inference of the pretrained model. It successfully reduces the time complexity by \textit{hundreds of times} in comparison with previous work following the traditional pipeline (Tab.~\ref{tab:efficiency}).
 
It is also worth mentioning that samples are selected based on the distance between extracted intermediate features. As a result, annotations are not required in the entire active learning pipeline. That is to say, our method is indeed \textit{unsupervised}, which relieves the troubles of assigning responsive annotators.

We conduct extensive experiments on different tasks, datasets, and networks. When compared with existing active learning methods following the traditional pipeline, our algorithm can beat almost all of them in performance with significantly advantageous generality and efficiency. Besides, our algorithm is very \textit{simple} and \textit{intuitive} with only hours of efforts for implementation. 

Our main contributions are summarized as follows.
\begin{itemize}
    \item For the first time, we propose a novel pipeline for active learning free from the repetitive model training and batch selection on each dataset. This pipeline adheres to three important principles: \textit{generality}, \textit{efficiency}, and \textit{non-supervision}, enabling an excellent time-efficiency for data selection close to random sampling.
    \item We design the first active learning algorithm, GEAL, following the new pipeline and satisfying all the three principles. It selects samples to fill in the annotation budget in one-shot based on our newly defined \textit{knowledge clusters} extracted from intermediate features.
    \item Our algorithm beats most traditional approaches in performance on popular tasks and datasets with significantly better efficiency. In particular, results of ablation study widely justify the design of our algorithm. These results verify the feasibility of our new pipeline and take a solid step towards general and efficient active learning.
\end{itemize}

\section{Related Work}
\label{sec:related}
\subsection{Active Learning} 

Active learning aims to choose the most suitable samples for annotation from a large unlabeled dataset so that model performance can be optimized with a limited annotation budget. Previous work can be divided into three categories: \textit{query-synthesizing methods}, \textit{stream-based methods} and \textit{pool-based methods}. \textit{Query-synthesizing methods} \cite{mahapatra2018efficient,mayer2020adversarial,zhu2017generative} synthesize samples based on the request of learners with a \textit{generative model}. This model is often difficult to train and highly specific to data domains. \textit{Stream-based methods} make a judgment on whether each sample should be labeled independently, which allows them to deal with sequential samples online. However, these methods usually require either a large dataset to train the selection policy \cite{fang2017learning,woodward2017active} or an online annotator to update the decision model \cite{mohamad2020online,narr2016stream}, both of which are unavailable in many cases.

Due to the difficulties of other two categories, most existing arts \cite{sinha2019variational,yoo2019learning,sener2017active,nguyen2004active,haussmann2020scalable,yuan2021multiple} follow the \textit{pool-based} protocol. Unlike \textit{stream-based methods}, it selects samples based on the ranking of the whole dataset. There exist two mainstream sampling strategies for pool-based methods \textit{i.e.} \textit{uncertainty} and \textit{diversity}. Uncertainty inside the model prediction reflects the difficulty of data samples, which is estimated by different heuristics such as probabilistic models \cite{gorriz2017cost,ebrahimi2019uncertainty}, entropy \cite{joshi2009multi,mackay1992information}, ensembles\cite{beluch2018power,li2013adaptive}, and loss function \cite{yoo2019learning,huang2021semi}. Unfortunately, many uncertainty-based methods cannot fit large datasets and deep models well \cite{sener2017active}. Alternatively, some other algorithms \cite{sener2017active,agarwal2020contextual,jain2016active} try to find the \textit{diverse} subset which well represents the entire data pool. Sener and Savarese \cite{sener2017active} theoretically justify that data selection is equivalent with \textit{k-Center} problem \cite{farahani2009facility} \textit{i.e.} minimizing the maximal distance between data points and selected samples. They implement it with Euclidean distance between high-dimensional features, which is challenged by \cite{sinha2019variational} and \cite{agarwal2020contextual}. Instead, Sinha et al. \cite{sinha2019variational} train a Variational Autoencoder (VAE) to represent the input, while Agarwal et al. \cite{agarwal2020contextual} apply the \textit{KL-divergence} between local region representations. However, all these methods, including some combinations of \textit{uncertainty} and \textit{diversity} \cite{kuo2018cost,wang2016cost}, rely on a small initial set to train the deep models for feature extraction. The size of initial set is insufficient to support a successful training process. As a compromise, they repetitively select data in batch, and train the model \textit{multiple times}, resulting in low efficiency.

Unlike all existing active learning research above, we propose a totally different pipeline in this paper, which selects samples on \textit{any} datasets in one-shot through \textit{a single-pass model inference}.

\subsection{Unsupervised Representation Learning} 
Currently, contrastive methods \cite{grill2020bootstrap,chen2020simple,he2020momentum,zhang2021self,xiao2020should,tian2020makes,patrick2020multi} have achieved great success in unsupervised representation learning, where models are trained to discriminate different images without using any explicit categories. Some early work \cite{chen2020simple,xie2020pointcontrast,he2020momentum,chen2020improved} implements it by bringing close representations from views of the same image (positive pairs) while splitting apart the representations from views of different images (negative pairs). By updating the parameters with moving average, Grill et al. \cite{grill2020bootstrap} successfully eliminate the negative pairs. Such a strategy inspires many following approaches \cite{xie2021propagate,li2020prototypical}. Recently, \cite{caron2021emerging} and \cite{chen2021empirical} explore the self-supervised learning of vision transformer \cite{dosovitskiy2020image}, where \cite{caron2021emerging} also shows that self-attention maps serve as a great indicator of region importance.

In this paper, we use the pretrained model \cite{caron2021emerging} to extract features on different datasets. As a result, we do not have to train models separately for each dataset like traditional active learning pipeline, ensuring the generality of our method.

\begin{figure}[tb!]
    \centering
    \includegraphics[width=0.95\textwidth]{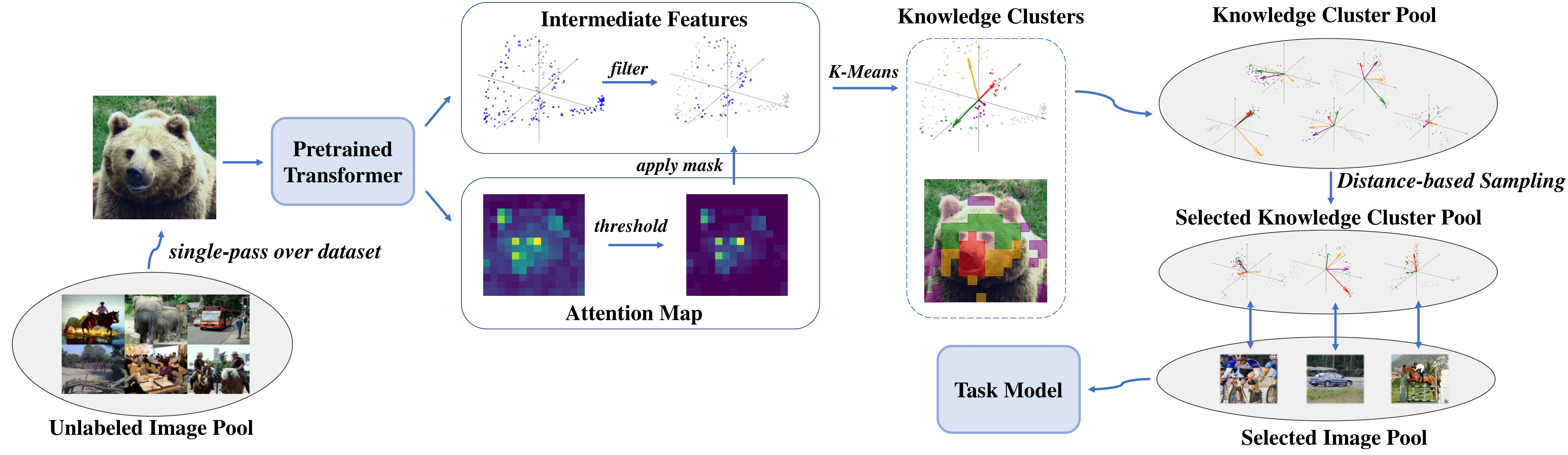}
    \caption{\textbf{Overview of Our Proposed GEAL:} Our method uses a general pretrained vision transformer to extract features from images. Knowledge clusters are derived from the intermediate features. Afterwards, we perform a distance-based sampling algorithm to select knowledge clusters as well as the associated images. These selected images are labeled for downstream task model training.}
    \label{fig:overview}
    \vspace{-10pt}
\end{figure}

\section{Methodology}
In this section, we will introduce our novel active learning method GEAL in detail. Firstly, in Sec.~\ref{sec:preliminary}, we show that trivial utilization of off-the-shelf global features extracted by a pretrained model fails in active learning. It spurs the necessity of designing a novel general and efficient active learning method. Then, we define a new notion called \textit{knowledge cluster} in Sec.~\ref{sec:knowledge}, which is important for GEAL. Afterwards, we elaborate on the sample selection strategy in Sec.~\ref{sec:sample}. The overview of our approach is illustrated in Fig.~\ref{fig:overview}. 

\begin{figure}[ht!]
    \centering
    \includegraphics[width=0.5\linewidth]{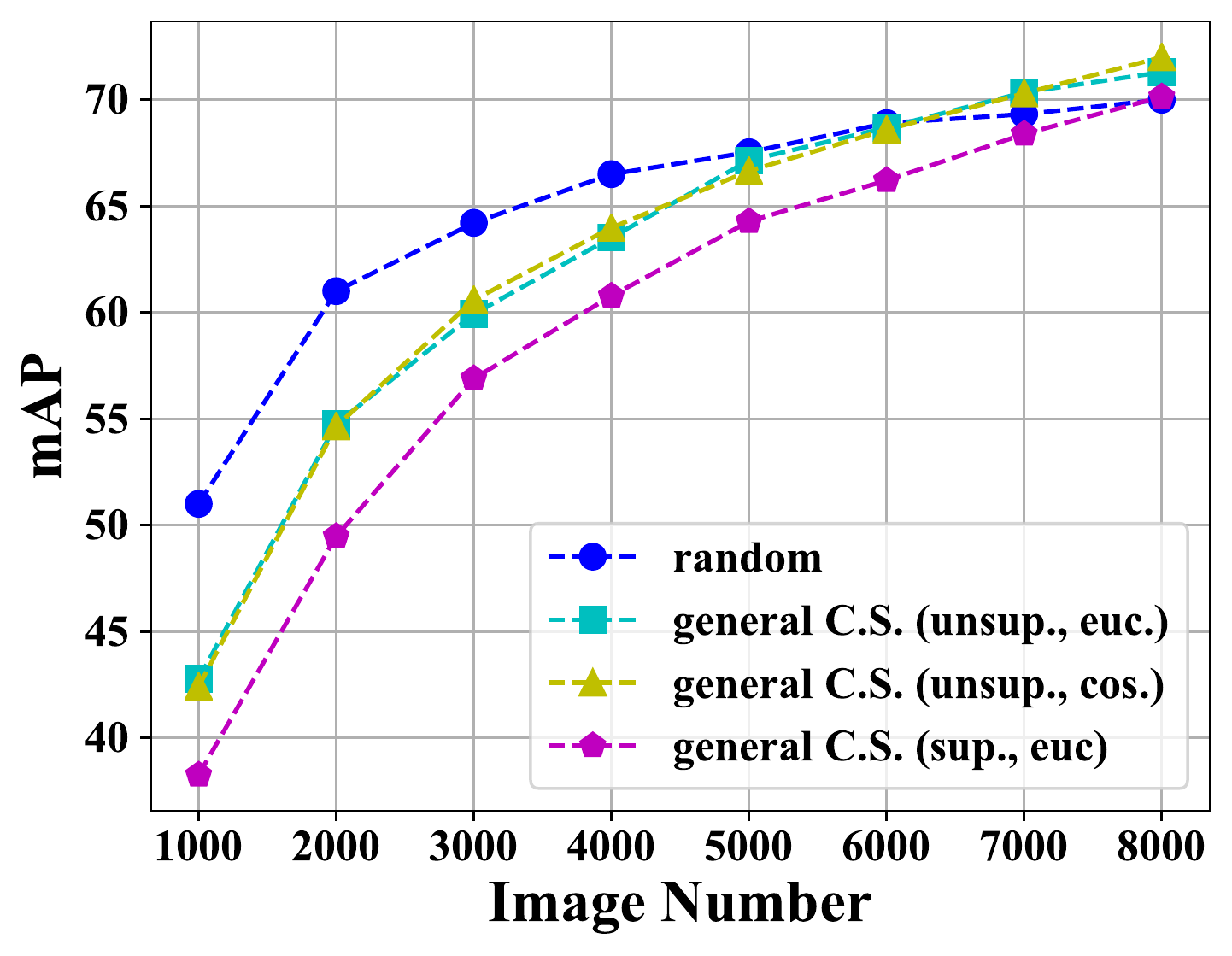}
    \caption{\textbf{Preliminary: Core-Set over Extracted Global Features:} \textit{C.S.} denotes Core-Set in this figure. We perform Core-Set algorithm over off-the-shelf global features extracted by DeiT-S model pretrained on ImageNet in both supervised or unsupervised manners. We follow \cite{sener2017active} to use Euclidean distance (\textit{euc.}) between features and try cosine distance (\textit{cos.}) as well. This setting causes a significant performance drop, especially when sampling ratio is low.}
    \label{fig:preliminary}
    \vspace{-10pt}
\end{figure}

\subsection{Off-the-Shelf Features for Active Learning}
\label{sec:preliminary}

Previous work in the field of active learning \cite{sener2017active,agarwal2020contextual,sinha2019variational} often selects representative samples based on the features extracted by deep models. These deep models are trained separately on each dataset. Here we take a step back and try to use off-the-shelf features instead, which are extracted by general models pretrained on a large-scale dataset. If it performs well, we can trivially improve the efficiency of active learning by eliminating the model training step on target datasets.

We conduct this experiment on object detection task over PASCAL VOC dataset \cite{everingham2010pascal} with SSD-300 model \cite{liu2016ssd}. Consistent with our following experiments, we apply DeiT-S \footnote{We follow the name in \cite{touvron2021training} throughout the paper. DeiT-S is called ViT-small in \cite{caron2021emerging}.} \cite{touvron2021training} model for feature extraction in the data selection process. The model is pretrained in either supervised or unsupervised (with DINO framework \cite{caron2021emerging}) manner on ImageNet dataset \cite{deng2009imagenet}. For active learning, we directly implement the classical Core-Set algorithm \cite{sener2017active} over the extracted global features, \text{i.e.} the last layer [CLS] token features. This algorithm selects different numbers of training samples, and we train object detection models over the selected subsets.

In Fig.~\ref{fig:preliminary}, we show results in comparison with random selection. Unfortunately, we find that active learning over off-the-shelf global features degrades the performance of the object detection task model, especially with relatively low sampling ratios. We consider two potential reasons for this failure:

\begin{itemize}
    \item \textbf{Complex scenes are hard to represent globally.} Images in PASCAL VOC dataset usually contain multiple objects including some very small ones, and object detection task relies heavily on local information. Thus, it is difficult for a global feature to represent all the useful details in the image.
    \item \textbf{K-Center tends to select simple scenes.} Global features of simple images are easily distant from each other if they contain distinct instances. In contrast, if there are multiple instances in complex images, information inside them is more likely to overlap with each other. As a result, in the feature space, to cover all the data samples with a small radius, K-Center algorithm prefers to select distinct simple images with few instances, which tend to be far from currently selected centers.
\end{itemize}

The above two concerns inspire our design of new active learning methods. We represent each image with dense features instead of a single global feature so that useful local information would not be ignored. Also, images are selected based on the distance between local \textit{knowledge clusters} instead of global features, which relieves the harmful preference for simple images.

\subsection{Knowledge Clusters inside Images}
\label{sec:knowledge}
In order to represent the useful local visual cues, we pay attention to intermediate features of pretrained networks. We consider the features of the $i$-th intermediate layer for image $I$ as $^I \mathbf{f}_i=\{^I f_i^r\in\mathbb{R}^{K_i}|r=1,2,\dots,H_iW_i\}$. Here $K_i$ is the feature dimension, while $H_i,W_i$ are the height and width of the feature map in layer $i$. $^{I}f^{r}_i\in\mathbb{R}^{K_i}$ is defined as the the feature corresponding to region $r$ in the feature map. For vision transformers \cite{dosovitskiy2020image,touvron2021training}, each patch corresponds to a region. For simplicity, we write $^I \mathbf{f}_i,^I f_i^r$ as $\mathbf{f}_i,f_i^r$ for image $I$ without ambiguity. In a deep layer, $f_i^r$ reflects the discriminative power of visual patterns inside the region $r$ \cite{li2021visualizing,raghu2021vision}. This makes sense in both convolutional models and vision transformers. 

However, it is still challenging to extract \textit{reliable} local information directly from $\mathbf{f}_i$ due to the following two concerns. \textbf{(i)} Given the low information density inside images, many regions are useless or even distracting for downstream tasks. \textbf{(ii)} Local features extracted by DNNs inevitably contain noise \cite{li2021visualizing}, \textit{i.e.}
\begin{equation}
    f_i^r=\hat{f}_i^r+\epsilon_i^r
\end{equation}
where $\hat{f}_i^r$ is the clean feature, and $\epsilon_i^r\overset{iid}{\sim}\mathcal{N}(\mathbf{0},\sigma^2 \mathbf{I}_{K_i})$ is noise \cite{li2021visualizing}. The noise exists because the model cannot perfectly represent the local information. 

For the first concern, the [CLS] token self-attention map of the transformer can serve as a natural indicator of regional importance even without dense supervision \cite{caron2021emerging}. For the $i$-th layer of image $I$, this self-attention map is denoted as $^I \mathbf{attn}_i=\{^I attn_i^r\in \mathbb{R}^+|r=1,\dots,H_iW_i\},\sum_{r=1}^{H_iW_i} {^I attn_i^r}=1$. We also write $^I \mathbf{attn}_i, ^I attn_i^r$ of image $I$ as $\mathbf{attn}_i,attn_i^r$ for simplicity. Since features in deep layers have stronger discriminative power \cite{li2021visualizing,raghu2021vision}, we obtain useful local features based on the last layer attention $\mathbf{attn}_{n}$ \footnote{Notice that the self-attention in the last layer is computed over the intermediate-features in the second last layer} of DeiT encoder \cite{dosovitskiy2020image,touvron2021training}. To be specific, we can filter the important regional features that jointly represent most information in the entire image, as shown in Eq.~\ref{eq:important}. 
\begin{equation}
    F(\mathbf{f}_{n-1})=\{f_{n-1}^r|r=1,2,\dots,t,\sum_{j=1}{^t attn_n^j} \leq \tau < \sum_{j=1}{^{t+1}attn_n^j}\}
    \label{eq:important}
    \vspace{0pt}
\end{equation}
where the regions $r=1,2,\dots,H_iW_i$ are sorted in the decreasing order of $attn_i^r$, and $\tau\in(0,1)$ is a hyper-parameter, meaning the maintained ratio of information represented by the filtered important features.

\begin{figure}[tb!]
    \centering
    \begin{subfigure}{0.31\textwidth}
        \centering
            \includegraphics[width=0.48\linewidth,height=0.48\linewidth]{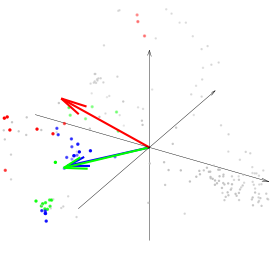}
            \includegraphics[width=0.48\linewidth,height=0.48\linewidth]{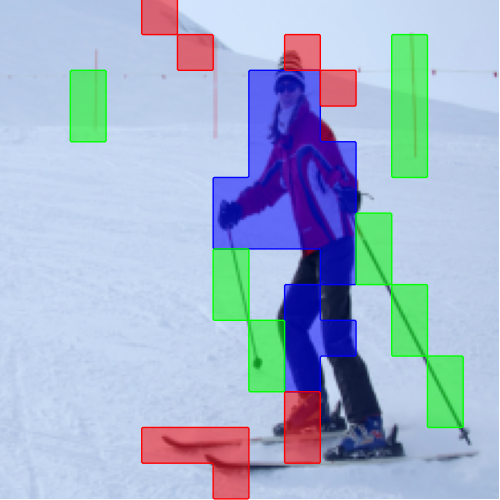}
    \end{subfigure}
    \hspace{0.01\textwidth}
    \begin{subfigure}{0.31\textwidth}
        \centering
        \includegraphics[width=0.48\linewidth,height=0.48\linewidth]{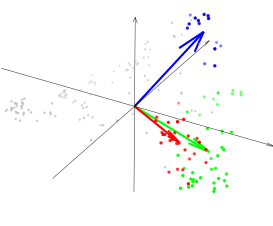}
        \includegraphics[width=0.48\linewidth,height=0.48\linewidth]{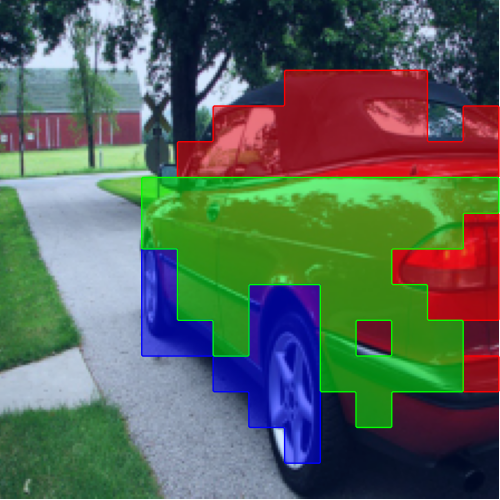}
    \end{subfigure}
    \hspace{0.01\textwidth}
    \begin{subfigure}{0.31\textwidth}
        \centering
        \includegraphics[width=0.48\linewidth,height=0.48\linewidth]{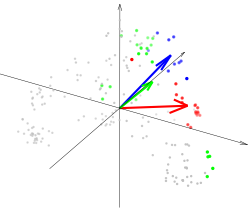}
        \includegraphics[width=0.48\linewidth,height=0.48\linewidth]{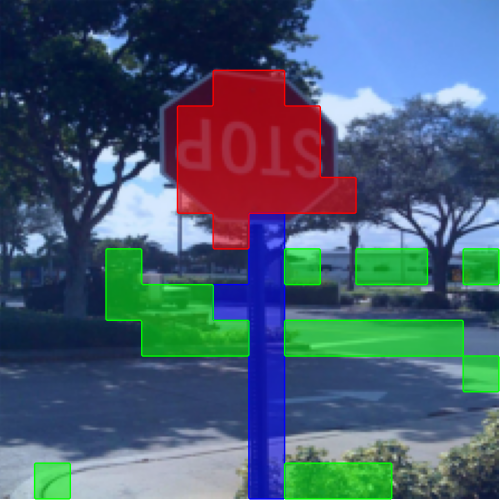}
    \end{subfigure}

    \caption{\textbf{Visualization of Knowledge Clusters:} Each two images are considered as a group. \textit{Left:} The filtered intermediate features of each image are grouped into different knowledge clusters. Gray features are eliminated in Eq.~\ref{eq:important}. Dimensions are reduced through PCA for visualization. \textit{Right:} Regions inside images can be associated with corresponding knowledge clusters.}
    \label{fig:knowledge}
    \vspace{-10pt}
\end{figure}

For the second concern, to eliminate the noise, we perform \textit{K-Means} clustering over the filtered intermediate features $F(\mathbf{f}_{n-1})$ to get $K$ centroids $^I \mathbf{c}=\{^I c_j|j=1,2,\dots,K\}$, where $K$ is a hyper-parameter.  Here $^I c_j,j=1,2,\dots,K$ are defined as \textit{knowledge clusters} inside the image $I$. We can derive $^I c_j$ in Eq.~\ref{eq:kmeans}. The noise can be greatly reduced if it obeys Gaussian distribution.
\begin{equation}
    \begin{aligned}
        ^I c_j=&\frac{1}{|C_j|}\sum_{r\in C_{j}}f_{n-1}^r
        =\frac{1}{|C_j|}\sum_{r\in C_{j}}\hat{f}_{n-1}^r + \frac{1}{|C_j|}\sum_{r\in C_{j}}\epsilon_{n-1}^r
        \approx \frac{1}{|C_j|}\sum_{r\in C_{j}}\hat{f}_{n-1}^r
    \end{aligned}
    \label{eq:kmeans}
    \vspace{0pt}
\end{equation}
where $f_{n-1}^r\in F(\mathbf{f}_{n-1}),r\in C_j$ are filtered features of image $I$ grouped into cluster $j$ through K-Means. The last approximation holds since $Var\left[\frac{1}{|C_j|}\sum_{r\in C_j}\epsilon_{n-1}^r\right]=\frac{1}{|C_j|}Var[\epsilon_{n-1}^r]=\frac{1}{|C_j|}\sigma^2\mathbf{I}_{K_i}$, which becomes closer to zero as $|C_j|$ grows. We visualize some examples of knowledge clusters $^Ic_j$ in Fig.~\ref{fig:knowledge}.

\subsection{Sample Selection Strategy for GEAL}
\label{sec:sample}
Our main target of data selection is to make the distributions of selected samples diverse and representative in the level of \textit{knowledge clusters} instead of images. This fine-grained strategy guarantees that our selected subset can cover rich local visual patterns, which are crucial for detection and segmentation tasks. 

Inspired by \cite{ash2019deep}, we transplant their distance-based sampling strategy to the \textit{knowledge cluster} level. The detailed algorithm is shown in Alg.~\ref{alg:selection}. In concise, it is similar to the K-Means++ seeding algorithm \cite{arthur2006k}. Given an unlabeled image pool $\mathcal{I}$, this process starts from randomly selecting an initial image $I_0$ \textit{i.e.} selecting all knowledge clusters $^{I_0}\mathbf{c}$ inside it. Then, we choose the next knowledge cluster $^Ic_j$ inside image $I$ with probability in proportion to its squared distances from the nearest already selected knowledge cluster (Eq.~\ref{eq:kseed}). 
\begin{equation}
    p(^Ic_j)\propto \min_{c\in s_{\mathcal{K}}}\left[d(^Ic_j,c)\right]^2
    \label{eq:kseed}
\end{equation}
where $s_{\mathcal{K}}$ is the pool of all the already selected knowledge clusters. When we choose a knowledge cluster $^Ic_j$, all the knowledge clusters $^I\mathbf{c}$ inside the image $I$ that contains $^Ic_j$ are put into the selected pool $s_{\mathcal{K}}$. We use cosine distance as the distance function $d(\cdot,\cdot)$ as analyzed in the \textit{supplementary materials}. This process continues until enough images have been selected. It is worth noting that the selection only requires knowledge clusters constructed from intermediate features offline beforehand. Consequently, only a \textit{single-pass} model inference \textit{without} any training or supervision is required in the entire active learning pipeline.

\begin{algorithm}[ht!]
    \caption{\textbf{Distance-based Sampling Strategy}}
    \label{alg:selection}
    \KwInput{all knowledge clusters $\mathbf{c}$, total annotation budget $b$}
    \KwOutput{selected image pool $s_{\mathcal{I}}$}
    Initialize $s_{\mathcal{I}}=\{I_0\}$\\
    \tcc{initialize the selected image pool with a random image $I_0$}
    
    Initialize $s_{\mathcal{K}}=\{^{I_0}c_j,j=1,\dots,K\}$\\ \tcc{initialize the selected knowledge cluster pool with all knowledge clusters inside $I_0$}
    
    \Repeat{$|s_{\mathcal{I}}|=b$}{
    Sample $^Ic_j$ with probability $p(^Ic_j)\propto \min_{c\in s_{\mathcal{K}}}\left[d(^Ic_j,c)\right]^2$\\
    \tcc{sample the next knowledge cluster $^Ic_j$ with probability in proportion to the squared distance}
    $s_{\mathcal{I}}=s_{\mathcal{I}}\cup \{I\}$\\
    \tcc{add image $I$ containing sampled $^Ic_j$ to selected image pool}
    $s_{\mathcal{K}}=s_{\mathcal{K}}\cup \{^Ic_k,k=1,\dots,K\}$\\
    \tcc{add all knowledge clusters inside image $I$ to selected knowledge cluster pool}
    }
\end{algorithm}

\section{Experiments}
We evaluate our proposed GEAL on the tasks of object detection (Sec.~\ref{sec:object}), image classification (Sec.~\ref{sec:image}), and semantic segmentation (Sec.~\ref{sec:semantic}). The results of GEAL are \textit{averaged over three independent selections with different random seeds}. Features are extracted by the same general pretrained model on all the datasets (Sec.~\ref{sec:pretrain}). We make some analysis of our proposed pipeline and method in Sec.~\ref{sec:analysis}. Finally, we examine the roles of different modules inside GEAL in Sec.~\ref{sec:ablation}. We refer readers to the \textit{supplementary materials} for more implementation details, analysis, and ablation studies.

\subsection{General Model for Feature Extraction}
\label{sec:pretrain}
We adopt DeiT-S \cite{touvron2021training} (path size 16$\times$16) pretrained with the unsupervised framework DINO \cite{caron2021emerging} on ImageNet \cite{deng2009imagenet} to extract features for data selection due to its verified effectiveness. The same pretrained model is used for all tasks. For each image $I$ in any datasets, knowledge clusters $^I \mathbf{c}$ are derived based on the last layer [CLS] token self-attention $^I \mathbf{attn}_n$ (average of multi-heads) and second last layer features $^I \mathbf{f}_{n-1}$ (Sec.~\ref{sec:knowledge}). We emphasize that this pretrained DeiT-S model is only applied to the data selection process. For the downstream tasks, we still train the convolutional task models from scratch in accordance with prior active learning work.

\subsection{Object Detection}
\label{sec:object}
\subsubsection{Dataset and Task Model} We carry out experiments on PASCAL VOC dataset \cite{everingham2010pascal}. In line with prior work \cite{agarwal2020contextual,yoo2019learning}, we combine the training and validation sets of PASCAL VOC 2007 and 2012 as the training data pool with $16,551$ images. The performance of task model is evaluated on PASCAL VOC 2007 test set using \textit{mAP} metric.
We follow previous work \cite{yoo2019learning,agarwal2020contextual} to use SSD-300 model \cite{liu2016ssd} with VGG-16 backbone \cite{simonyan2014very}, which would be trained on the selected samples. This model reaches $77.43$ mAP with $100\%$ training data.

\begin{figure}[t]
    \centering
    \begin{minipage}[t]{0.47\textwidth}
        \centering
        \includegraphics[width=\textwidth]{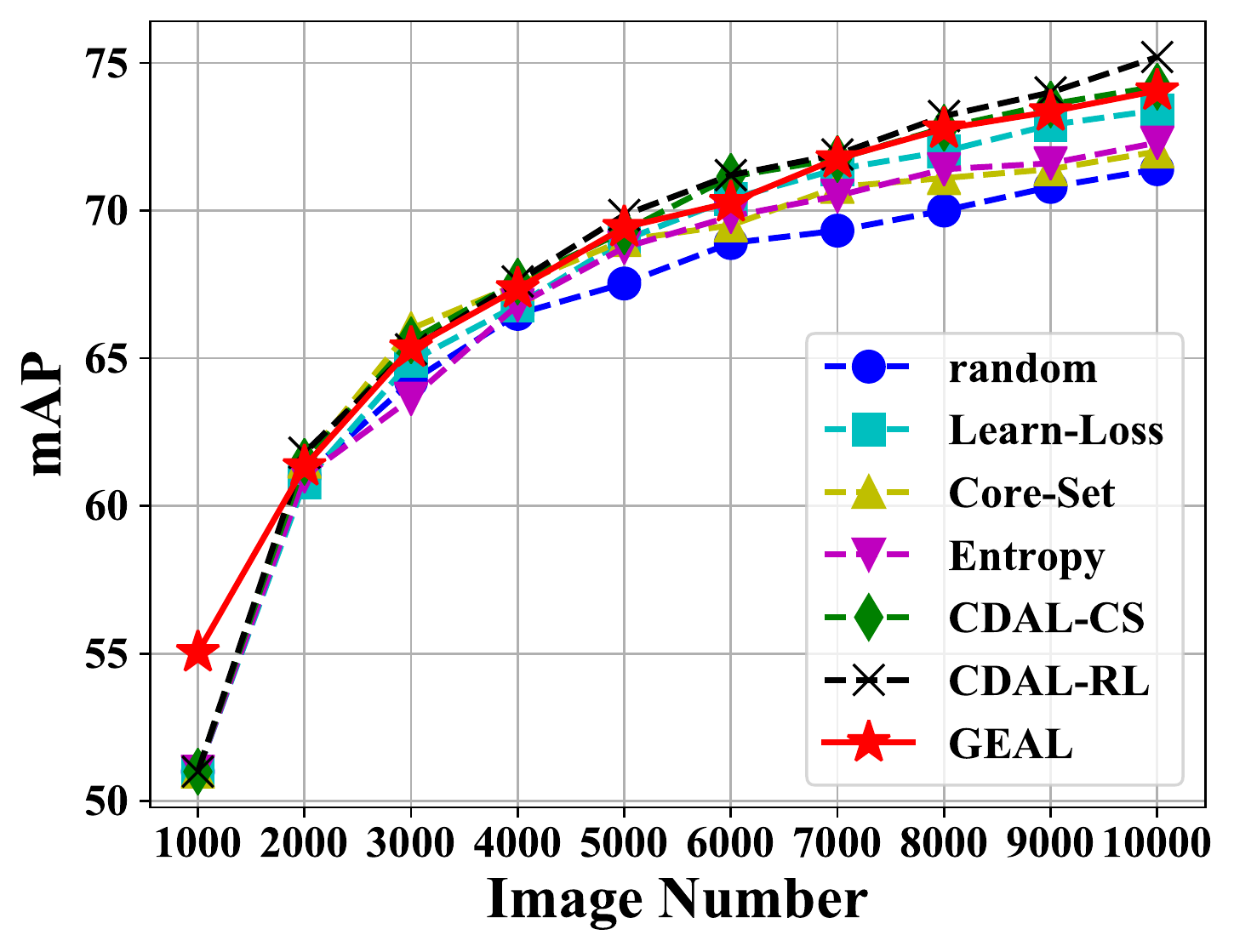}
        \caption{\textbf{Results on Object Detection:} The mAP on 100\% training data is 77.43. Results are averaged over three independent runs.}
        \label{fig:object}
        \vspace{-10pt}
    \end{minipage}
    \hspace{0.04\textwidth}
    \begin{minipage}[t]{0.47\textwidth}
        \centering
        \includegraphics[width=\textwidth]{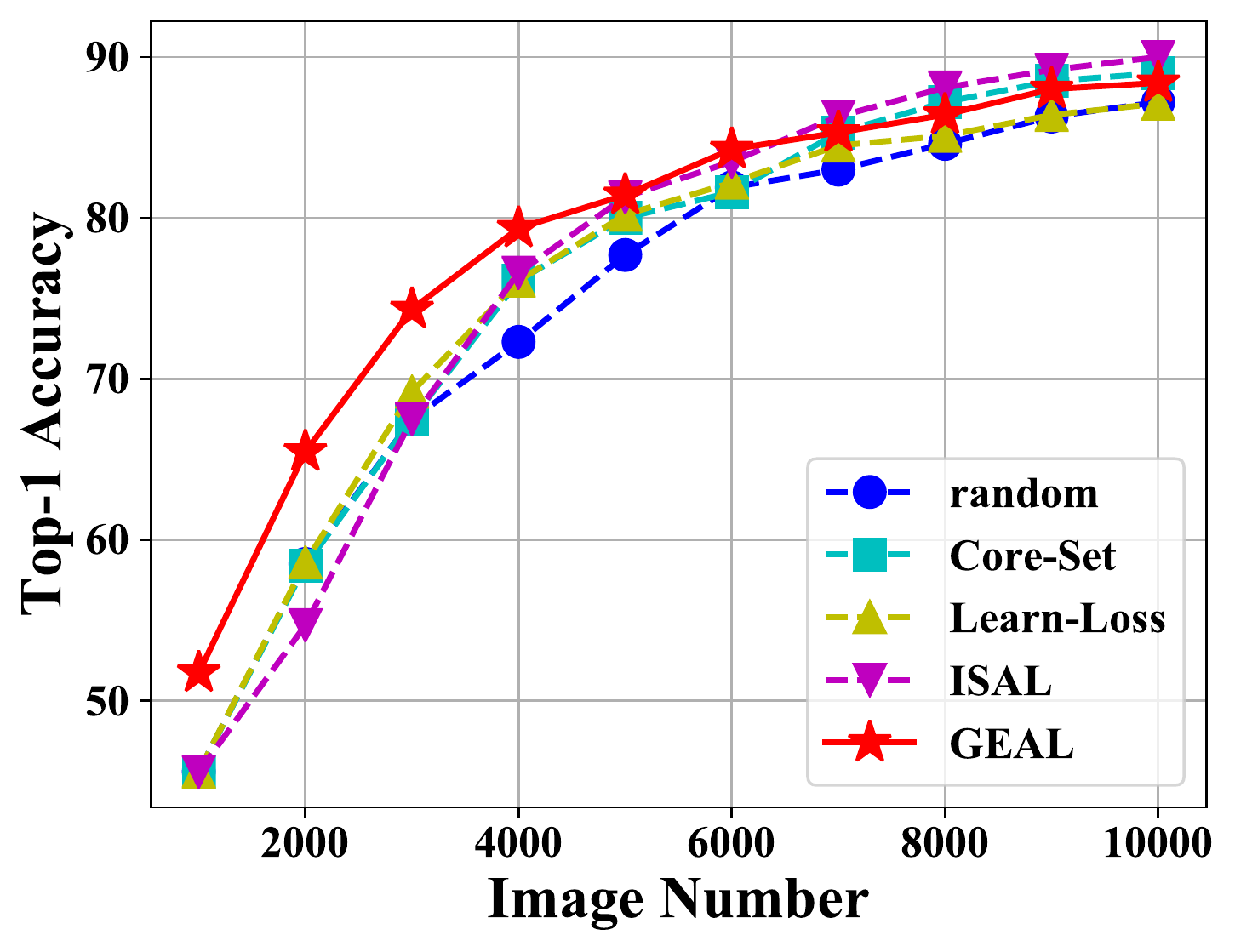}
        \caption{\textbf{Results on Image Classification:} The Top-1 Accuracy on 100\% training data is 93.02\%. Results are averaged over three independent runs.}
        \label{fig:cls_res}
         \vspace{-10pt}
    \end{minipage}
\end{figure}

\subsubsection{Results and Comparison} 
We show the performance of active learning methods on object detection with different sampling ratios in Fig.~\ref{fig:object}. The performance of GEAL is compared with state-of-the-arts following traditional pipeline. For fairness, we only include task-agnostic methods instead of those designed specific for object detection \cite{yuan2021multiple,choi2021active} which should naturally perform better.

Results show that GEAL outperforms most traditional pipeline methods. It also remains competitive with the best one, CDAL-RL \cite{agarwal2020contextual}, which needs extra reinforcement learning training. Besides, it is worth emphasizing that all these previous methods require repetitive model training and data selection on each target dataset separately, while GEAL can select all samples in one-shot without additional training. Given our extreme superiority to all the counterparts in generality and efficiency (Tab.~\ref{tab:efficiency}), this performance is already very impressive.

\subsection{Image Classification}
\label{sec:image}
\subsubsection{Dataset and Task Model} 
We use CIFAR-10 dataset \cite{krizhevsky2009learning} and ResNet-18 \cite{he2016deep} model in line with prior work \cite{liu2021influence,yoo2019learning}. CIFAR-10 contains 60,000 images with scale 32$\times$32 belonging to 10 categories (50,000 for training and 10,000 for test). We report the results using \textit{Top-1 Accuracy} metric. The model reaches $93.02\%$ Top-1 Accuracy with $100\%$ training data.

\subsubsection{Results} 
We demonstrate the results of active learning in Fig.~\ref{fig:cls_res}. Our performance is compared with state-of-the-arts following the traditional pipeline as well. With most sampling ratios, GEAL beats all its counterparts. Of particular interest, when the sampling ratio is low, our advantage is quite significant, enabling the model to achieve a satisfactory performance earlier. 

\begin{figure}[t]
    \centering
    \begin{minipage}[t]{0.47\textwidth}
        \centering
        \includegraphics[width=\textwidth]{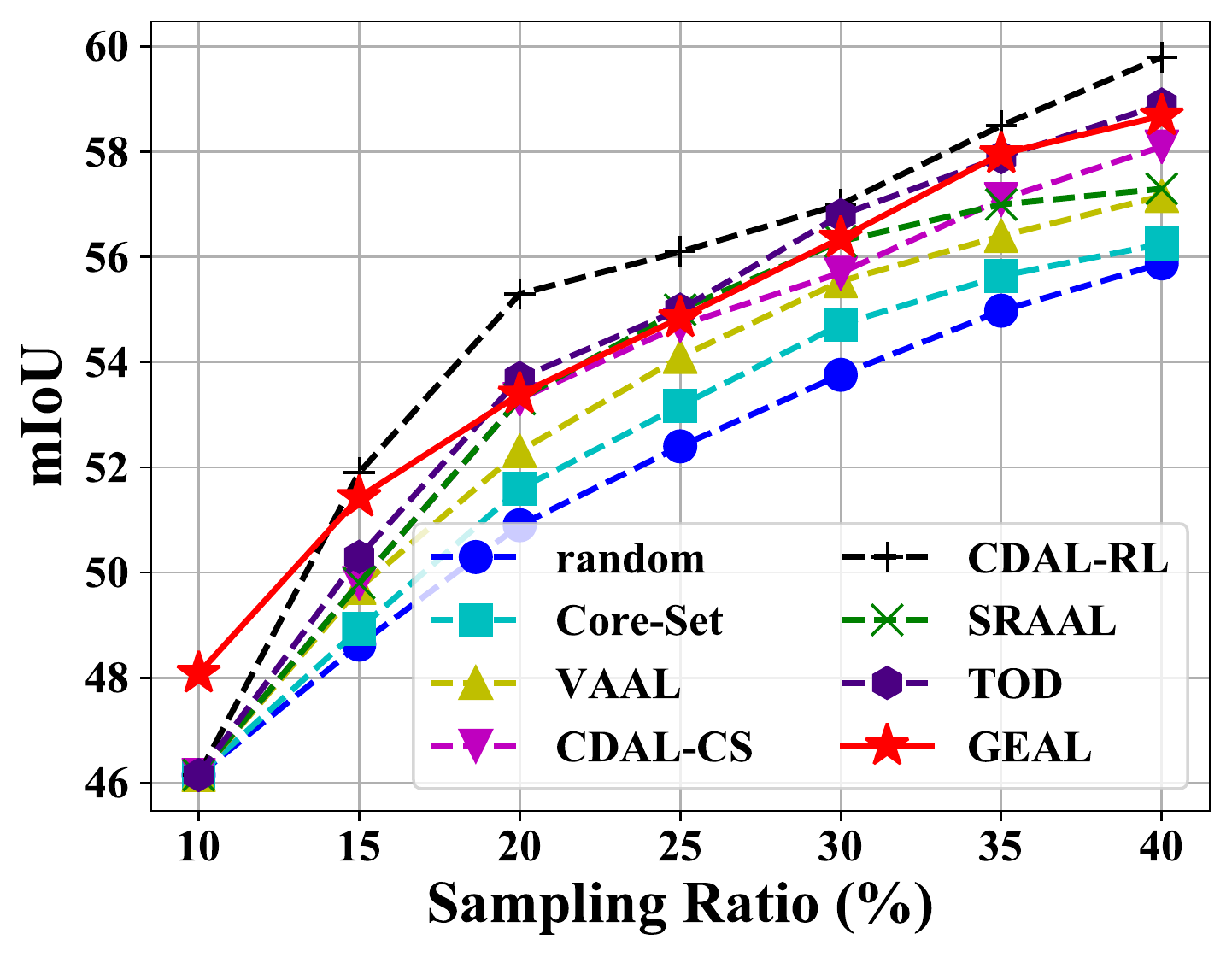}
        \caption{\textbf{Results on Semantic Segmentation:} The mIoU on 100\% training data is 62.95. Results are averaged over three independent runs.}
        \label{fig:seg}
        \vspace{-10pt}
    \end{minipage}
    \hspace{0.04\textwidth}
    \begin{minipage}[t]{0.47\textwidth}
        \centering
        \includegraphics[width=\textwidth]{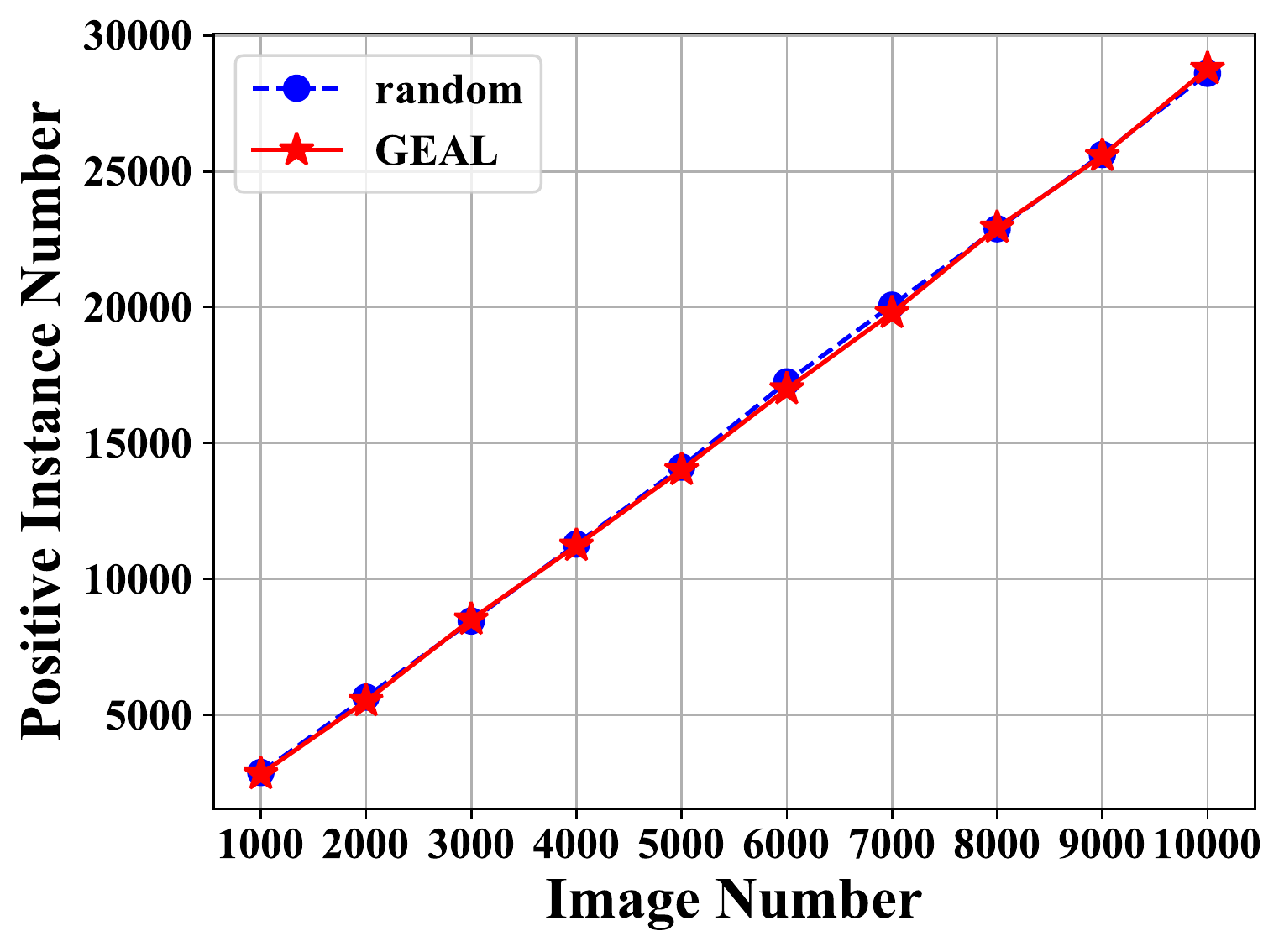}
        \caption{\textbf{Selected Instance Number:} We count the number of positive instances inside selected PASCAL VOC images.}
        \label{fig:count}
        \vspace{-10pt}
    \end{minipage}
\end{figure}

\subsection{Semantic Segmentation}
\label{sec:semantic}
\subsubsection{Dataset and Task Model} 
We use the common Cityscapes \cite{cordts2016cityscapes} dataset for semantic segmentation, same with prior work. This dataset is composed of 3,475 frames with pixel-level annotation of 19 object classes. We report the result using \textit{mIoU} metric. We follow previous active learning research to apply DRN \cite{yu2017dilated} model for this task. It reaches $62.95$ mIoU with $100\%$ training data.

\subsubsection{Results}
We compare the performance of GEAL with traditional active learning methods \footnote{For a fair comparison of data selection quality, task models are trained supervisedly on selected samples \textit{i.e.} we do not apply the semi-supervised setting in TOD \cite{huang2021semi}.} in Fig.~\ref{fig:seg}. Given the domain gap between the pretraining dataset (\textit{i.e.} ImageNet) and Cityscapes, it is quite challenging to apply our general active learning method to this task. However, similar to object detection, GEAL is only slightly outperformed by CDAL-RL \cite{agarwal2020contextual}, which requires additional reinforcement learning. With much higher efficiency, GEAL still beats most traditional pipeline methods in performance.

\subsection{Analysis}
\label{sec:analysis}

\subsubsection{Instance Number} We count the sum of instance numbers inside the selected images of PASCAL VOC in Fig.~\ref{fig:count}. Interestingly, although GEAL achieves much better performance than random selection, the number of selected instances are similar to random selection. This counter-intuitive fact reveals that GEAL could find the most informative and diverse high-quality instances even without supervision. For one thing, it avoids the annotation of complex images, reducing labor and expense. For another, this phenomenon motivates us to rethink the correlation between model performance and annotation quantity.


\subsubsection{Time Efficiency Analysis} Time efficiency of active learning methods is crucial for its practical use. Tab.~\ref{tab:efficiency} shows the comparison between GEAL and other existing counterparts. The estimation is conducted on PASCAL VOC dataset where $8,000$ samples should be chosen. We follow prior papers \cite{agarwal2020contextual,sener2017active,yoo2019learning} to use SSD-300 as task model with VGG-16 backbone (same as Sec.~\ref{sec:object}). The time is estimated on a single NVIDIA TITAN RTX GPU. For our GEAL, since it directly utilizes the publicly available pretrained model \textit{instead of} training models separately for each dataset, only the feature extraction, knowledge cluster construction, and data selection time should be considered, \textit{i.e.} 155 seconds in total regardless of the task model.  In contrast, for other work following the traditional pipeline, networks are trained repetitively on each dataset. Compared to the training, their model inference and data selection time can be ignored for convenience. We follow \cite{yoo2019learning,agarwal2020contextual} to set their initial set size and batch selection budget both as $1k$. In this case, their model should be trained for $7$ times over subsets of size $1k\sim 7k$ to select 8,000 samples. For convenience, we optimistically estimate their training efficiency the same as object detector SSD-300, which requires about 42 hours in total. In this case, we ignore the time cost of their extra modules \textit{e.g.} the loss estimation module in Learn-Loss \cite{yoo2019learning} and reinforcement learning in CDAL-RL \cite{agarwal2020contextual}. These previous methods also need to wait the oracle for ground-truth labels after selecting each batch of data. Based on the above information, our method is \textbf{at least 975x faster} than prior work. 

\subsubsection{Elimination of Initial Set} Unlike prior active learning methods, our GEAL following the new pipeline is free of a random initial set. It allows for great practical use.  Firstly, GEAL is the \textit{only} method that brings performance gain in the lowest sampling ratio, verified in our experiments on three tasks. This is beneficial in practice when the annotation budget is extremely low. Besides, GEAL would not suffer from the imbalanced initial set. As discussed in \cite{sinha2019variational}, low-quality initial sets would hurt the performance of prior active learning work significantly, \textit{e.g.} the lack of some categories in the initial set. However, it is difficult to ensure the high-quality of initial sets in practice if we have limited knowledge of the entire pool, \textit{e.g.} we may not know the category number. Therefore, with no need of the initial set, GEAL could have better compatibility in practical situations.

\subsubsection{Introduction of Pretrained Model} The pretrained model is a new module introduced in our proposed pipeline (Fig.~\ref{fig:our_pipeline}). This introduction is inevitable to satisfy the three principles of the new pipeline. Since the pretraining is not designed specific for the active learning task, directly using a publicly available model would not lead to extra time-cost or expense. According to Sec.~\ref{sec:preliminary}, it is non-trivial to improve the efficiency of active learning with a pretrained model, which reflects the value of GEAL. We will further show that our great performance does not come from the introduction of pretrained model in Sec.~\ref{sec:ablation}.

\begin{figure}[tb]
    \centering
    \begin{minipage}[b]{0.55\textwidth}
    \centering
    \tabcaption{\textbf{Module Contribution:} We discuss the contribution of each module inside GEAL. KC means knowledge cluster. Experiments are conducted on PASCAL VOC.} 
    \label{tab:module}
    \resizebox{\textwidth}{!}{
    \begin{tabular}{cccccc}
        \toprule
         \multirow{2}{*}{\textbf{Feature}} & \multirow{2}{*}{\textbf{Ratio}} & \multirow{2}{*}{\textbf{Select}} & 
         \multicolumn{3}{c}{\textbf{Image Number}}\\
         & & & 3k & 5k & 7k\\
         \midrule
         global & \ding{55} & FDS & 60.59 & 66.65 & 70.30\\
         KC & \ding{55} & FDS & 63.87 & 68.07 & 71.30\\
         KC & $\tau=0.5$ & FDS & 64.62 & 69.00 & 71.74\\
         KC & $\tau=0.5$ & Prob. & \textbf{65.35} & \textbf{69.43} & \textbf{71.76}\\
         \midrule
         \multicolumn{3}{c}{\textit{random sampling}} & 64.21 & 67.53 & 69.32\\
         \bottomrule
    \end{tabular}
    }
  \end{minipage}
   \hspace{0.01\textwidth}
  \begin{minipage}[t]{0.42\textwidth}
    \centering
    \includegraphics[width=\linewidth]{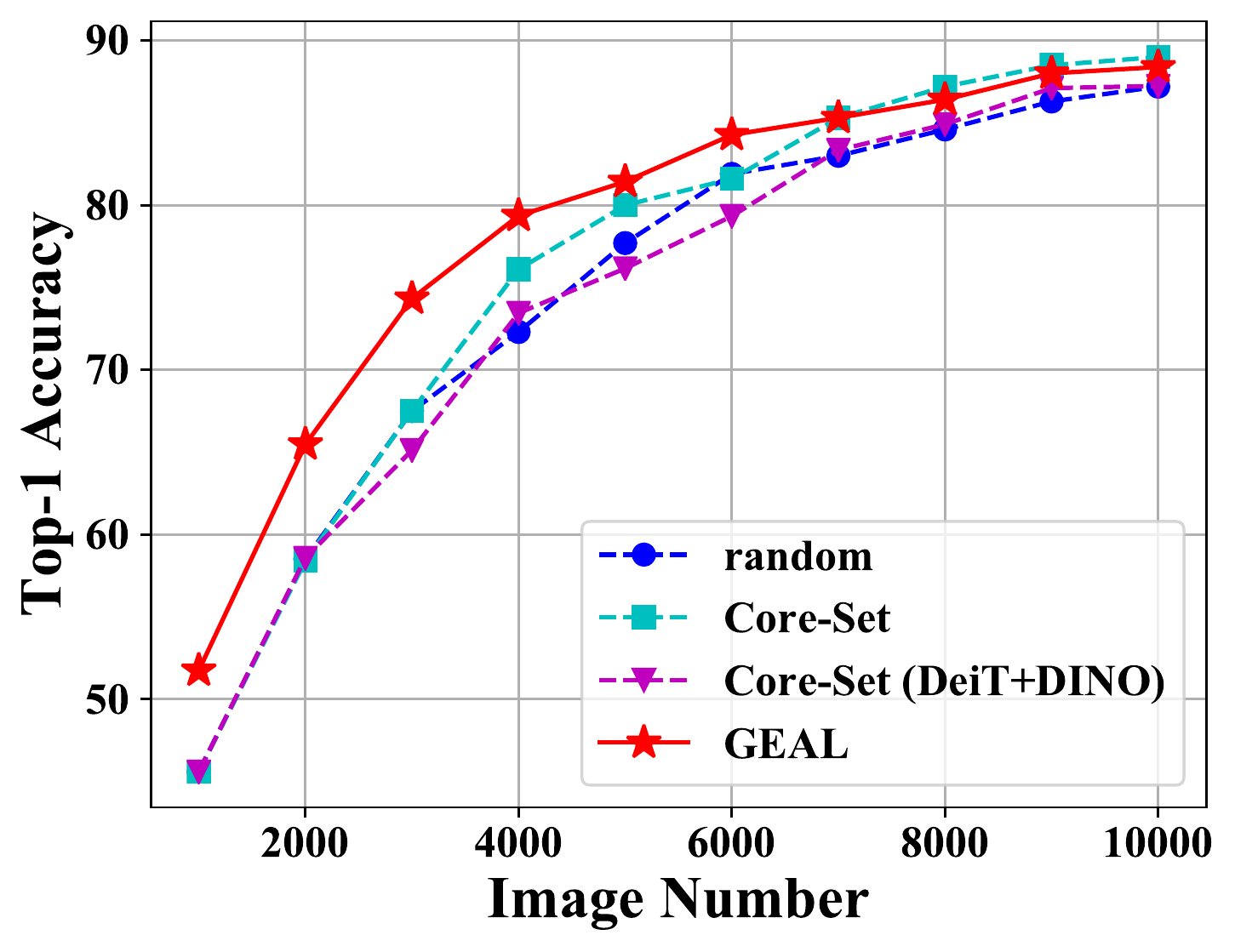}
    \caption{\textbf{Role of the Pretrained DeiT-S:} We modify Core-Set by selecting data with DeiT-S pretrained with DINO. Experiments are conducted on CIFAR-10.}
    \label{fig:pretrain}
      \vspace{-10pt}
  \end{minipage}
\end{figure}

\subsection{Ablation Study}
\label{sec:ablation}
We give a deep insight into different parts of our method. Firstly, we analyze the contribution of each module inside GEAL to the final performance step by step. Then, the role of the pretrained DeiT-S model is also analyzed. 

\subsubsection{Each Module Contribution}
Starting from the failure case, \textit{i.e.} general Core-Set (unsup., cos.) in Fig.~\ref{fig:preliminary}, modules of GEAL are added to it one-by-one. Tab.~\ref{tab:module} demonstrates the step-by-step performance improvement. This experiment is conducted on PASCAL VOC in the same setting with Sec.~\ref{sec:object}. The following three modules are analyzed. More quantitative analysis of hyper-parameters is available in the \textit{supplementary materials}.
\begin{itemize}
    \item \textbf{Feature Extraction Manner:} In Sec.~\ref{sec:preliminary}, the global feature of [CLS] token is directly used. We replace it with the knowledge clusters defined in Eq.~\ref{eq:kmeans}.
    \item \textbf{Attention Ratio:} We apply the attention ratio in Eq.~\ref{eq:important} to filter the local features, discarding other noisy information.
    \item \textbf{Selection Strategy:} Apart from the distance-based probability sampling in Eq.~\ref{eq:kseed}, we consider the alternative farthest-distance-sampling (FDS) \textit{w.r.t.} knowledge clusters, which is theoretically justified in \cite{sener2017active}. It chooses the next knowledge cluster farthest from the nearest selected one as $^Ic_j=arg\max_{^Ic_j}\min_{c\in s_{\mathcal{K}}}d(^Ic_j,c)$.
\end{itemize}

The failure case of general Core-Set is shown in the first line of Tab.~\ref{tab:module}. Then, we extract features by constructing knowledge clusters ($K=5$) without applying the attention ratio in the second line. It improves notably compared with the first line because the knowledge clusters can represent useful local information important for object detection. However, it is only slightly better than random selection since the knowledge clusters are dominated by local noisy information in this stage. To this end, we apply attention ratio $\tau=0.5$ in the third line of the table, and the performance gets improved again. Finally, the FDS selection strategy is replaced by the distance-based probability sampling in Eq.~\ref{eq:kseed}. It provides extra performance gain because it would select fewer outliers, making these selected samples more representative.

\subsubsection{Role of Pretrained Model} There is a concern that the performance gain of GEAL comes from the great representation ability and transferability of the pretrained vision transformer for data selection instead of our designed method. About this, we conduct an ablation study on CIFAR-10 in the same setting as Sec.~\ref{sec:image}. We equip Core-Set \cite{sener2017active} method with the same network for data selection, \textit{i.e.} DeiT-S \cite{touvron2021training} pretrained with DINO \cite{caron2021emerging}. During the data selection period, pretrained DeiT-S is finetuned supervisedly and used to select data samples with Core-Set algorithm. When the selection is done, we still train ResNet-18 \cite{he2016deep} model over the selected samples from scratch. In Fig.~\ref{fig:pretrain}, this pretrained DeiT-S damages the performance of Core-Set. A potential explanation comes from the different diversity in the representation spaces of pretrained DeiT and from-scratch ResNet. Consequently, the samples selected by DeiT-S are not suitable for the from-scratch training of ResNet-18 model. This reflects that our performance gain does not come from the use of pretrained DeiT-S. Instead, the proposed GEAL plays an important role.

\section{Conclusion and Limitations}
The main goal of this paper is to propose a novel active learning pipeline with three important principles: generality, efficiency, and non-supervision. We verified its feasibility by designing the first method GEAL following this new pipeline. Challenging the necessity of repetitively training models on each dataset, we extracted features with a general pretrained model. Through a single-pass model inference, the knowledge clusters were constructed based on the intermediate features, over which a distance-based selection strategy was performed to find the most diverse and informative data samples in one-shot. We demonstrated promising results on different tasks including object detection, image classification, and semantic segmentation. Our method outperformed most traditional pipeline counterparts with a near-a-thousand-times superiority of efficiency.

We realized that GEAL cannot beat all the active learning approaches following the traditional pipeline in current stage because of the absence of training and supervision on the target datasets. Nevertheless, this direction is still worth exploring due to its superior generality and efficiency, as well as notable potential in practical use, enabling data selection with a speed close to random sampling. We believe this paper takes the important and solid first step, which would inspire further work towards general and efficient active learning.

\clearpage
%
%
\bibliographystyle{splncs04}
\bibliography{arxiv}

\clearpage
\appendix

In this appendix, we include some additional ablation studies in Sec.~\ref{sec:ablation} to justify our design of GEAL. Then, the stability of GEAL results is analyzed in Sec.~\ref{sec:stability}. Afterwards, we provide the pseudo-code for the first part of GEAL, knowledge cluster extraction, in Sec.~\ref{sec:pseudo} while the pseudo-code of the second part is already in Alg.~1 of our main paper. Finally, implementation details of our experiments are explained in Sec.~\ref{sec:implementation}. 

\begin{figure}[t]
    \centering
    \begin{minipage}[t]{0.47\textwidth}
        \centering
        \includegraphics[width=\textwidth]{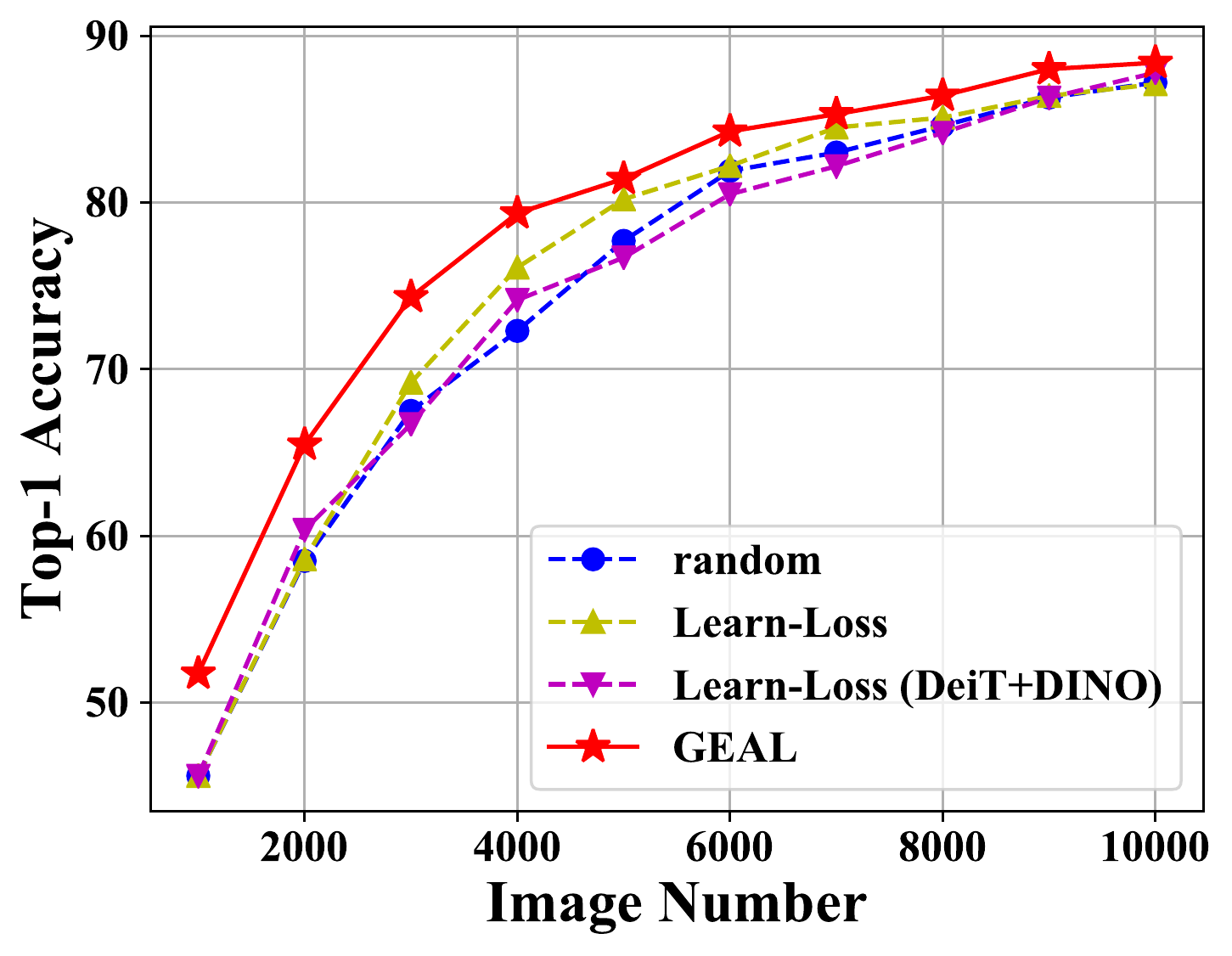}
        \caption{\textbf{Learn-Loss with Pretrained Model:} We equip Learn-Loss \cite{yoo2019learning} active learning method with DeiT-S model \cite{touvron2021training} pretrained with DINO \cite{caron2021emerging} for data selection. Experiments are conducted on CIFAR10 image classification task.}
        \label{fig:learnloss_role}
    \end{minipage}
    \hspace{0.04\textwidth}
    \begin{minipage}[t]{0.47\textwidth}
        \centering
        \includegraphics[width=\textwidth]{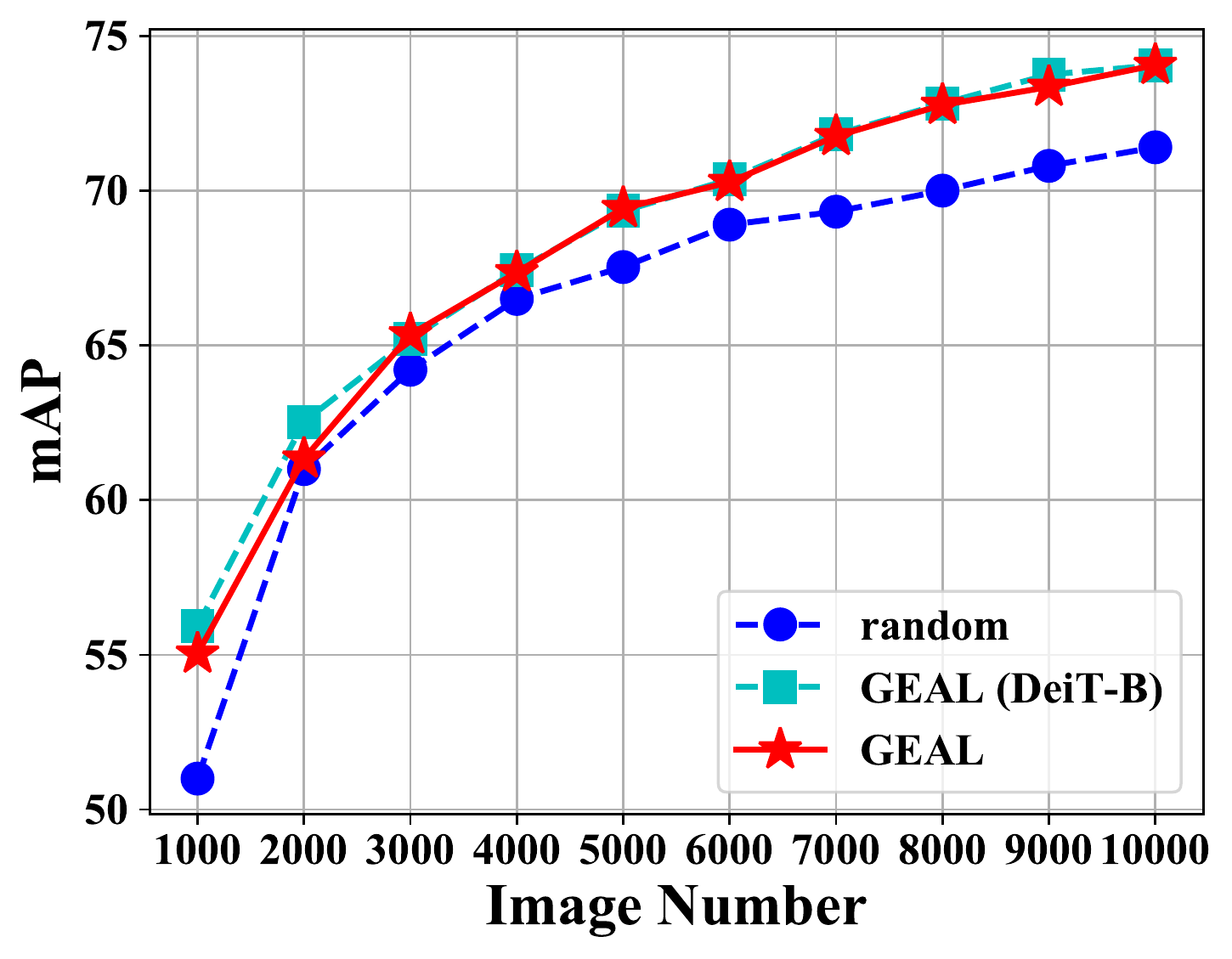}
        \caption{\textbf{DeiT-B as Feature Extractor:} We replace pretrained DeiT-S (22M parameters) in our GEAL with stronger DeiT-B (86M parameters) as feature extractor for data selection. Experiments are conducted on PASCAL VOC object detection task.}
        \label{fig:deitb}
    \end{minipage}
\end{figure}

\section{Additional Ablation Studies}
\label{sec:ablation}
\subsection{Additional Analysis of Pretrained Model Role}
Similar with Sec.~4.6 of our main paper, we conduct another supplementary experiment on image classification task to show that traditional active learning algorithms cannot benefit from a pretrained feature extractor. Just like what we do in Sec.~4.6 for Core-Set \cite{sener2017active}, the Learn-Loss \cite{yoo2019learning} active learning algorithm is equipped  with a DeiT-S \cite{touvron2021training} model pretrained with DINO \cite{caron2021emerging} to select data samples on CIFAR10 \cite{krizhevsky2009learning}. In the data selection process, the pretrained DeiT-S is finetuned to select data in batch with Learn-Loss algorithm. When the data has been selected, we still train the ResNet-18 model \cite{he2016deep} from scratch to evaluate the data selection quality same as Sec.~4.3 of our main paper.

Results in Fig.~\ref{fig:learnloss_role} show that the pretrained model hurts the active learning performance of Learn-Loss on CIFAR10 with most sampling ratios. This phenomenon is consistent with our results for Core-Set algorithm \cite{sener2017active} in Sec.~4.6 of our main paper. Therefore, We conclude that the pretrained DeiT-S model cannot directly boost traditional active learning algorithms, so the superior performance of GEAL does not come from the pretrained model. Instead, our designed modules in GEAL are actually important.

\subsection{Stronger Pretrained Model}
GEAL makes use of DeiT-S \cite{touvron2021training} model pretrained with DINO framework \cite{caron2021emerging} for data selection. In this part, we replace DeiT-S (22M parameters) with larger DeiT-B (86M parameters) as the feature extractor. We conduct the experiments on object detection task, where SSD-300 \cite{liu2016ssd} task model with VGG-16 backbone \cite{simonyan2014very} is trained on the selected subset of PASCAL VOC dataset \cite{everingham2010pascal} in the same setting as Sec.~4.2 of our main paper. Interestingly, in Fig.~\ref{fig:deitb}, despite the stronger representation ability of DeiT-B, we find the active learning performance is similar with the original GEAL with DeiT-S. Therefore, the ability and transferability of pretrained model only have limited influence on our active learning performance. Instead, our designed GEAL method is the key factor to decide the performance of active learning.

\begin{table}[tb!]
\caption{\textbf{Sensitivity to Hyperparameters:} $\tau,K,d(\cdot,\cdot)$ separately denote the attention ratio, knowledge cluster number, and distance function. Experiments are conducted on PASCAL VOC object detection task.}
\label{tab:sensitivity}
\centering
\setlength{\tabcolsep}{2mm}{
    \begin{tabular}{ccccccc}
        \toprule
        \multirow{2}{*}{$\boldsymbol{\tau}$} & \multirow{2}{*}{$\mathbf{K}$} & \multirow{2}{*}{$\mathbf{d(\cdot,\cdot)}$} &
        \multirow{2}{*}{\textbf{Pretraining}} &
        \multicolumn{3}{c}{\textbf{Image Number}}\\
         & & & & $3k$ & $5k$ & $7k$\\
         \midrule
        0.3 & \multirow{3}{*}{5} & \multirow{3}{*}{cos.} &  \multirow{3}{*}{\textit{unsupervised} \cite{caron2021emerging}} & 64.90 & 69.00 & 71.36\\
        0.5 &  & & & \textbf{65.35} & \textbf{69.43} & 71.76\\
        0.7 &  & & & 64.48 & 68.53& 71.58 \\
        \midrule
        \multirow{2}{*}{0.5} & 1 & \multirow{2}{*}{cos.} & \multirow{2}{*}{\textit{unsupervised} \cite{caron2021emerging}} & 64.90 & 69.01 & 71.02 \\
         & 10 & & &65.17 & 69.41 &71.52\\
        \midrule
        0.5 & 5 & euc. & \textit{unsupervised} \cite{caron2021emerging} & 65.28 & 69.08 & \textbf{71.89}\\
        \midrule
        0.5 & 5 & cos. & \textit{supervised} \cite{touvron2021training} & 64.23 & 68.71 & 71.44\\
        \bottomrule
    \end{tabular}
}
\end{table}

\subsection{Sensitivity to Hyperparameters}
In this part, we analyze the sensitivity of our GEAL to some hyperparameters including the attention ratio $\tau$ (Eq.~2 of our main paper), the knowledge cluster number $K$ (Eq.~3 of our main paper, j=1,2,$\dots$, K), the distance function $d(\cdot,\cdot)$ (Eq.~4 of our main paper), and pretraining manner for the general model. Experiments are conducted on object detection task, where samples are selected from PASCAL VOC dataset in the same setting as Sec.~4.2 of our main paper. Results are shown in Tab.~\ref{tab:sensitivity}.

\subsubsection{Attention Ratio $\tau$} Attention ratio $\tau$ notably affects the final performance of GEAL. Too low ratios lead to the ignorance of some crucial local visual patterns, while too high ratios introduce some harmful noisy patterns to the knowledge clusters. Thus, a moderate attention ratio plays an important role in the high performance of GEAL.

\subsubsection{Knowledge Cluster Number $K$} When $K=1$, the performance is hurt since knowledge clusters degrade to global features in this case. If $K>1$, GEAL is not very sensitive to different knowledge cluster numbers.

\subsubsection{Distance Function $d(\cdot,\cdot)$} Cosine distance leads to slightly better performance than Euclidean distance. This result reflects that the directions of intermediate features are important for the diversity of local visual patterns.

\subsubsection{Pretraining Manner} Instead of using the unsupervised pretraining framework DINO \cite{caron2021emerging}, we try the DeiT-S
model \cite{touvron2021training} pretrained in a supervised manner on ImageNet \cite{krizhevsky2012imagenet}. Results show the drop of performance in this case. We think that the supervised pretraining introduces some biases to the pretrained model.

\begin{figure}[t]
    \centering
    \includegraphics[width=0.6\textwidth]{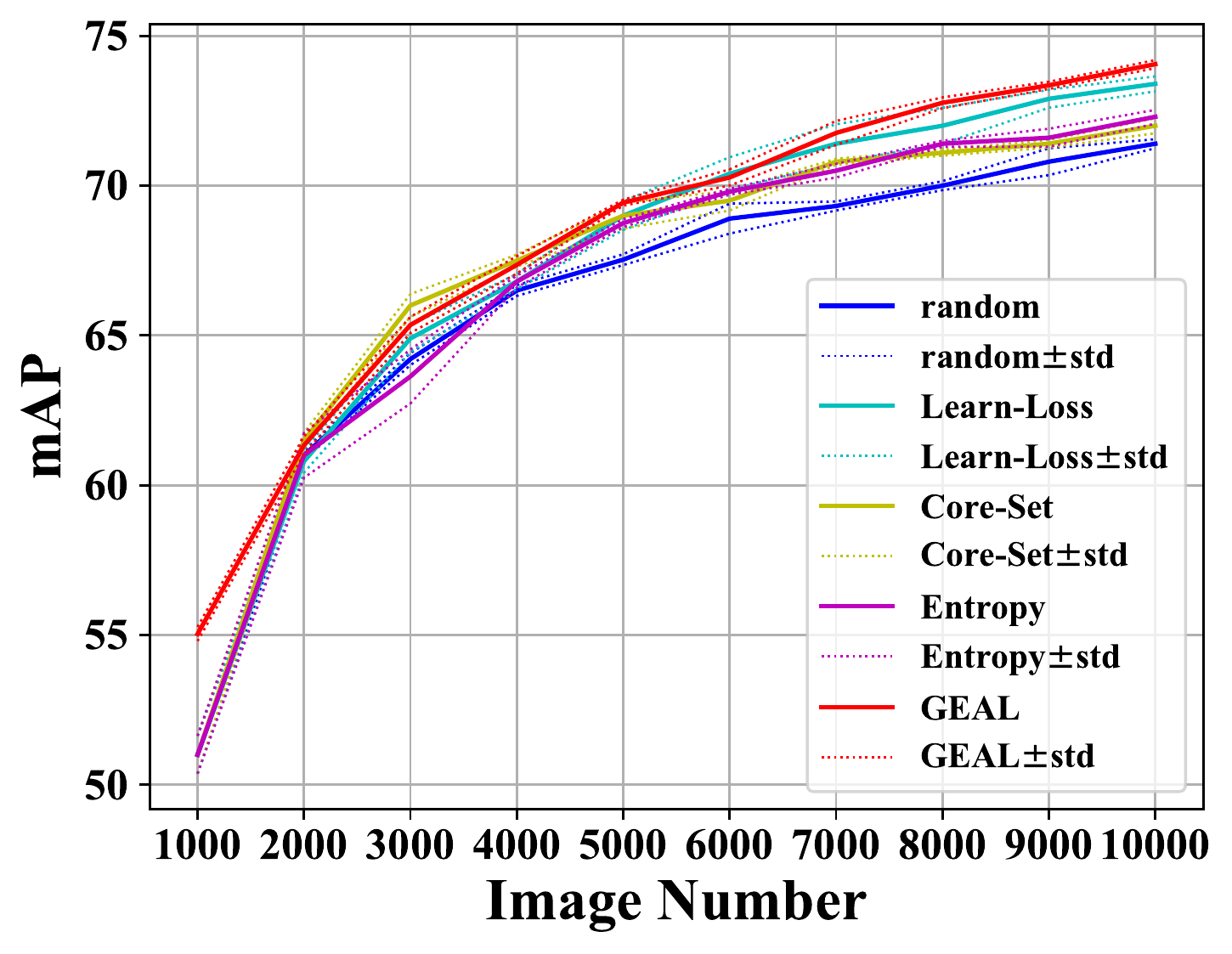}
    \caption{\textbf{Stability of Active Learning Results:} Experiments are conducted on PASCAL VOC object detection task. Our mean and standard deviation are calculated over three independent trials.}
    \label{fig:std}
\end{figure}

\section{Additional Analysis: Stability of Results}
\label{sec:stability}
On PASCAL VOC dataset \cite{everingham2010pascal} of object detection task, we compare the stability of GEAL and some traditional pipeline methods \footnote{CDAL paper \cite{agarwal2020contextual} does not report the standard deviations, so this method is not taken into account in the comparison}. In Fig.~\ref{fig:std}, the standard deviation of the performance of GEAL is smaller than those of traditional methods. Traditional methods rely heavily on the random initial set to select data, which leads to high randomness inside the selection. In contrast, GEAL eliminates this initial set, making our data selection more stable.

\section{Pseudo-Code for Knowledge Cluster Extraction}
\label{sec:pseudo}
In this section, we provide the pseudo-code for the first part of GEAL, \textit{i.e.} knowledge cluster extraction (Sec.~3.2 of the main paper). The pseudo-code for the second part (\textit{i.e.} data selection) is already shown in Alg.~1 of our main paper. Given a pretrained network $g$ and unlabeled image pool $\mathcal{I}$, we aim to obtain the knowledge clusters $^I\mathbf{c}$ for each image $I\in\mathcal{I}$. The detailed algorithm is explained in Alg.~\ref{alg:knowledge}.

\begin{algorithm}[t!]
    \caption{\textbf{Knowledge Cluster Extraction}}
    \label{alg:knowledge}
    \KwInput{Pretrained deep neural network $g$, unlabeled image pool $\mathcal{I}$, attention ratio $\tau$, centroid number $K$}
    \KwOutput{Knowledge clusters $^I \mathbf{c},I\in\mathcal{I}$}
    \For{$I\in\mathcal{I}$}{
        $^I \mathbf{attn}_n,^I\mathbf{f}_{n-1}=g(I)$\\
        \tcc{extract the last layer attention and second last layer intermediate features from the network}
        \tcc{only a single-pass model inference is required in this process}
        Sort $^I attn_n^r,^I f_{n-1}^r,r=1,\dots,H_nW_n$ in the decreasing order of attention\\
        \tcc{regions $r=1,\dots,H_nW_n$ should be arranged in the decreasing order of their attention $^I attn_n^r$, where $H_n,W_n$ is the height and width of the last layer feature map. Each patch corresponds to a region in vision transformers.}
        $F(^I\mathbf{f}_{n-1})=\{^If_{n-1}^r|r=1,\dots,t,$ $\sum_{j=1}^t {^I attn_n^j} \leq \tau < \sum_{j=1}^{t+1}{^I attn_n^j}\}$\\
        \tcc{filter important local features based on the sorted attention (Eq.~2 of our main paper)}
        
        $^I\mathbf{c}=KMeans(F(^I\mathbf{f}_{n-1}), K)$\\
        \tcc{get $K$ centroids $^I c_j,j=1,\dots,K$ by conducting K-Means over $F(^I\mathbf{f}_{n-1})$ for each image $I$}
    }
\end{algorithm}

\section{Implementation Details}
\label{sec:implementation}

\subsection{Object Detection Implementation}
\subsubsection{Implementation of GEAL} We set attention ratio $\tau=0.5$ and knowledge cluster number $K=5$. The input images are resized to $224\times 224$ when fed into the pretrained DeiT-S \cite{touvron2021training} model in the data selection process.

\subsubsection{Implementation of Task Model} The implementation of task model is same as previous active learning research \cite{yoo2019learning,agarwal2020contextual}. The SSD-300 model \cite{liu2016ssd} with VGG-16 \cite{simonyan2014very} backbone is adopted for this experiment. The model is implemented based on mmdetection \footnote{https://github.com/open-mmlab/mmdetection}. We follow \cite{yoo2019learning,agarwal2020contextual} to train the model for $300$ epochs with batch size $32$ using SGD optimizer (momentum $0.9$). The initial learning rate is $0.001$, which decays to $0.0001$ after $240$ epochs.

\subsection{Semantic Segmentation Implementation}
\subsubsection{Implementation of GEAL} Same as object detection, we set attention ratio $\tau=0.5$ and knowledge cluster number number $K=5$. The input images are resized to $448\times 224$ in line with their original aspect ratios when fed into the pretrained DeiT-S \cite{touvron2021training} model in the data selection process.

\subsubsection{Implementation of Task Model} We follow prior active learning work \cite{sinha2019variational,huang2021semi} to apply DRN \cite{yu2017dilated} model \footnote{https://github.com/fyu/drn} for semantic segmentation task. The model is trained for 50 epochs with batch size 8 and learning rate 5e-4 using Adam optimizer \cite{kingma2014adam}. 

\subsection{Image Classification Implementation}
\subsubsection{Implementation of GEAL} We follow previous tasks to set attention ratio $\tau=0.5$. Since image classification depends less on local information, we directly set the centroid number $K=1$. The input images are resized to $224\times 224$ when fed into the pretrained DeiT-S \cite{touvron2021training} model in the data selection process.

\subsubsection{Implementation of Task Model} We follow \cite{yoo2019learning,liu2021influence} to use ResNet-18 \cite{he2016deep} classification model in this task, which is implemented based on mmclassification \footnote{https://github.com/open-mmlab/mmclassification}. The model is trained for 200 epochs with batch size 128 using SGD optimizer (momentum 0.9, weight decay 5e-4). The initial learning rate is 0.1, which decays to 0.01 after 160 epochs. We apply standard data augmentation to the training including 32×32 size random crop from 36×36 zero-padded images and random horizontal flip.

\end{document}